%% file: ijcai17.tex
\documentclass{article}
\usepackage{ijcai17}

\usepackage{times}

\usepackage{helvet}
\usepackage{courier}
\usepackage{amsmath}
\usepackage{amssymb}
\usepackage{url}

\usepackage{color}
\usepackage{amsmath}
\usepackage{booktabs}
\usepackage{multirow}
\usepackage{graphicx}
\usepackage{subfigure}
\usepackage{epsfig}
\usepackage{enumerate}
\DeclareMathOperator*{\argmax}{argmax}   
\title{Long-Term Memory Networks for Question Answering}

\author{Fenglong Ma$^{*1}$, Radha Chitta$^{*2}$, Saurabh Kataria$^{*3}$\\
\textbf{Jing Zhou}$^{*2}$\textbf{, Palghat Ramesh}$^{4}$\textbf{, Tong Sun}\thanks{Work carried out while at PARC, a Xerox Company.}$^{5}$\textbf{, Jing Gao}$^{1}$\\
$^{1}$SUNY Buffalo, $^{2}$Conduent Labs US, 
$^{3}$LinkedIn\\
$^{4}$PARC, 
$^{5}$United Technologies Research Center\\
\{fenglong, jing\}@buffalo.edu, \{radha.chitta, jing.zhou\}@conduent.edu\\
saurabh.cse05@gmail.com, palghat.ramesh@parc.com, sunt@utrc.utc.com
}

\begin{document}

\maketitle

\input{0-abstract}
\input{1-introduction}

\input{3-relatedwork}
\input{2-model}
\input{4-experiments}

\input{5-conclusion}

\begin{small}
\bibliographystyle{named}
\bibliography{ijcai17}
\end{small}

\end{document}

%% file: 0-abstract.tex
\begin{abstract}
Question answering is an important and difficult task in the natural language 
processing domain, because many basic natural language processing tasks can be cast 
into a question answering task. 
Several deep neural network architectures have been developed recently, 
which employ memory and inference components to memorize and reason over text information, and generate answers to questions.  
However, a major drawback of many such models is that they are capable of only generating single-word answers. In addition, they require large amount of training data to generate accurate answers.  
In this paper, we introduce the Long-Term 
Memory Network (LTMN), which incorporates both an external memory module and a Long Short-Term Memory (LSTM) 
module to comprehend the input data and generate multi-word answers.
The LTMN model can be trained end-to-end using back-propagation and requires minimal supervision.
We test our model on two synthetic data sets (based on Facebook's bAbI data set) and the real-world Stanford question 
answering data set, and show that it can achieve 
state-of-the-art performance.
\end{abstract}

%% file: 1-introduction.tex
\section{Introduction}
Question answering (QA), a challenging problem which requires an ability to understand and analyze the 
given unstructured text, is one of the core tasks in natural language understanding and processing. 
Many problems in natural language processing, such as reading comprehension, 
machine translation, entity recognition, sentiment analysis, and dialogue generation, can be cast 
as question answering problems. 

Traditional question answering approaches can be categorized as: 
(i) IR-based question answering \cite{pacsca2003open} where the question is 
 formulated as a search query, and a short text 
segment is found on the Web or similar corpus for the answer;
(ii) Knowledge-based question answering \cite{green1961baseball,berant2013semantic}, 
 which aims to answer a natural language question by mapping it to a semantic query over 
 a database.

The traditional approaches are simple query-based techniques. It is difficult to establish the relationships between the sentences in the input text, and derive a meaningful representation 
of the information within the text using these traditional question-answering systems. 


\begin{figure}[!htb]\small
 \centering
\begin{minipage}[h]{0.45\textwidth}
\fbox{
\parbox{\textwidth}{
{\tt{
1: Burrel's innovative design, which combined the low production cost of an 
Apple II with the computing power of Lisa's CPU, the Motorola 68K, received the 
attention of Steve Jobs, co-founder of Apple.\\
2: Realizing that the Macintosh was more marketable than the Lisa, he began to 
focus his attention on the project.\\
3: Raskin left the team in 1981 over a personality conflict with Jobs.\\
4: {\color{blue}Why did Raskin leave the Apple team in 1981?}	{\color{red}over 
a personality conflict with Jobs}\\
5: Team member Andy Hertzfeld said that the final Macintosh design is closer to 
Jobs' ideas than Raskin's.\\
6: {\color{blue}According to Andy Hertzfeld, whose idea is the final Mac design 
closer to?}	{\color{red}Jobs	}\\
7: After hearing of the pioneering GUI technology being developed at Xerox PARC, 
Jobs had negotiated a visit to see the Xerox Alto computer and its Smalltalk 
development tools in exchange for Apple stock options.\\
8: {\color{blue}What did Steve Jobs offer Xerox to visit and see their latest 
technology?} {\color{red}Apple stock options}
}}
}
}
\end{minipage}
\caption{Example of a question answering task.}\label{fig:example}
\end{figure}

Figure \ref{fig:example} shows an example of question answering task. The 
sentences in black are facts that may be relevant to the questions, 
questions are in blue, and the correct answers are in red. In order to correctly answer
the question ``\textit{What did Steve Jobs offer Xerox to visit and see their latest 
technology?}'', the model should have the ability to recognize that the sentence 
``\textit{After hearing of the pioneering GUI technology being developed at 
Xerox PARC, Jobs had negotiated a visit to see the Xerox Alto computer and its 
Smalltalk development tools in exchange for Apple stock options.}'' is a supporting fact 
and extract the relevant portion of the supporting fact to form the answer. In addition, the model should have the ability to 
memorize all the facts that have been presented to it until the current time, and deduce the 
answer.

The authors of \cite{weston2015memory} proposed a new class of 
learning models named Memory Networks (MemNN), which use a long-term memory 
component to store information and an inference component for reasoning. 
\cite{DBLP:conf/icml/KumarIOIBGZPS16} proposed the Dynamic Memory Network (DMN) 
for general question answering tasks, which processes input sentences and 
questions, forms episodic memories, and generates answers. These two approaches 
are \textbf{strongly supervised}, i.e., only the supporting facts (factoids) are fed to the 
model as inputs for training the model for each type of question. For example, when training the model with the question in the fourth line of Figure \ref{fig:example}, 
strongly supervised methods only use the sentence in line 3 as input.
Thus, these methods require a large amount of training data.

To tackle this issue, \cite{sukhbaatar2015end} introduced a \textbf{weakly 
supervised} approach called End-to-End Memory Network (MemN2N), which uses all the sentences 
that have appeared before this question. For the above example, the inputs are the sentences from line 1 to line 3 when training for the question in the fourth line. 
MemN2N is 
trained end-to-end and uses an attention mechanism to calculate the matching 
probabilities between the input sentences and questions. The sentences which match the question
with high probability are used as the factoids for answering the question.

However, this model is capable of generating only \emph{single-word} answers. 
For example, the answer of the question ``\textit{According to Andy 
Hertzfeld, whose idea is the final Mac design closer to?}'' in Figure 
\ref{fig:example} is only one word ``\textit{Jobs}''. Since the answers of many 
questions contain \textbf{multiple words} (for instance, the question labeled $4$ 
in Figure~\ref{fig:example}), this model cannot be directly applied to the 
general question answering tasks. 

Recurrent neural networks comprising Long Short Term Memory Units have been employed to generate multi-word text in the literature~\cite{graves2013generating,sutskever2014sequence}. However, simple LSTM based recurrent neural networks do not perform well on the question-answering task due to the lack of an external memory component which can memorize and contextualize the facts. 
We present a more sophisticated recurrent neural network 
architecture, named Long-Term Memory Network (LTMN), which combines the best aspects of end-to-end memory networks and LSTM based recurrent neural networks to address the challenges faced by the currently available neural network architectures for question-answering. Specifically, it first 
embeds the input 
sentences (initially encoded 
using a distributed representation learning mechanism such as paragraph vectors \cite{le2014distributed}) in a continuous space, 
and stores them in memory. It then matches the sentences with the questions,
also embedded into the same space, 
by performing multiple passes through the memory, 
to obtain the factoids which are relevant to each question. 
These factoids are then employed to generate the 
first word of the answer, which is then input to an LSTM unit. The LSTM 
unit is used to generate the subsequent words in the answer. The proposed LTMN model can be trained 
end-to-end, requires minimal supervision during training (i.e., weakly supervised), and 
generates multiple words answers. 
Experimental results on two synthetic datasets and one real world dataset show that the proposed model outperforms the state-of-the-art approaches.


In summary, the contributions of this paper are as follows:

\begin{itemize}
\item We propose an effective neural network architecture for general question answering, i.e. for generating multi-word answers for questions. Our architecture combines the best aspects of MemN2N and LSTM and can be trained end-to-end.
\item The proposed architecture employs distributed representation learning techniques (e.g. paragraph2vec) to learn vector representations for sentences or factoids, questions and words, as well as their relationships. The learned embeddings contribute to the accuracy of the answers generated by the proposed architecture.
\item We generate a new synthetic dataset with multiple word answers based on Facebook's bAbI dataset \cite{weston2016towards}. We call this the multi-word answer bAbI dataset.  
\item We test the proposed architecture on two synthetic datasets (the single-word answer bAbI dataset and the multi-word answer bAbI dataset), and the real-world Stanford question answering dataset~\cite{rajpurkar2016squad}. The results clearly demonstrate the advantages of the proposed architecture for question answering.
\end{itemize}

%% file: 3-relatedwork.tex
\section{Related Work}
In this section, we review literature closely related to question answering, 
particularly focusing on models using memory networks to generate answers. 
\subsection{Question Answering}
Traditional question answering approaches mainly include two categories: IR-based 
\cite{pacsca2003open} and Knowledge-based question answering 
\cite{green1961baseball,berant2013semantic}. IR-based question answering systems 
use information retrieval techniques to extract information (i.e., answers) from 
documents. These methods first process questions, i.e., detect named entities 
in questions, and then predict answer types, such as cities' names or person's 
names. After recognizing answer types, these approaches generate queries, and extract answers from the web using the generated 
queries. These approaches are easy, but they ignore the semantics between questions 
and answers.

Knowledge-based question answering systems 
\cite{DBLP:conf/uai/ZettlemoyerC05,berant2014semantic,zhang2016joint} consider the semantics 
and use existing knowledge bases, such as Freebase \cite{bollacker2008freebase} 
and DBpedia \cite{Bizer:2009:DCP:1640541.1640848}. They cast the question 
answering task as that of finding one of the missing arguments 
in a triple. Most of knowledge-based question answering approaches use neural 
networks, dependency trees and knowledge bases \cite{bordes2012joint} or 
sentences \cite{iyyer2014neural}.

Using traditional question answering approaches, it is difficult to establish the relationship between sentences in the input text, and thereby identify the relevance of the different sentences to the question. Of late, several neural network architectures with memories have been proposed to solve 
this challenging problem.

\subsection{Memory Networks}
Several deep neural network models use memory architectures 
\cite{sukhbaatar2015end,DBLP:conf/icml/KumarIOIBGZPS16,weston2015memory,graves2014neural,joulin2015inferring,mozer1993connectionist}
and attention mechanisms for image captioning \cite{you2016image}, machine comprehension \cite{wange2016mploying} and healthcare data mining \cite{ma2017dipole,suo2017amia}. We focus on the models using memory networks for 
natural language question answering.

Memory networks (MemNN), proposed in \cite{weston2015memory}, first introduced the 
concept of an external memory component for natural language question answering. They are strongly 
supervised, i.e., they are trained with only the supporting facts for each question.
The supporting input sentences are embedded in memory, and the response is generated from these facts 
by scoring all the words in the vocabulary in correlation with the facts. This scoring function is learnt during the training 
process and employed during the testing phase. MemNN are capable of producing only single-word answers, due to this response 
generation mechanism. In addition, MemNN cannot be trained end-to-end.

The authors of \cite{DBLP:conf/icml/KumarIOIBGZPS16} improve over MemNN by introducing an end-to-end trainable network called Dynamic Memory Networks (DMN). DMN have four modules: input module, question 
module, episodic memory module and answer module. The input module encodes raw 
text inputs into distributed vector representations using a gated recurrent 
network (GRU) \cite{cho2014properties}. The question module 
similarly encodes the question using a recurrent neural network. 
The sentences and question representations are fed to the episodic memory module, 
which chooses the sentences to focus on using the attention mechanism. It iteratively 
produces a memory vector, representing all the relevant information, which is then used by the answer module 
to generate the answer using a GRU. However, DMN are also strongly supervised like MemNN, thereby requiring a large amount of training data. 

End-to-End Memory Networks (MemN2N) \cite{sukhbaatar2015end} 
first encode sentences into continuous vector 
representations, then use a soft attention mechanism to calculate matching 
probabilities between sentences and questions and find the most relevant facts, and finally 
generate responses using the vocabulary from these facts. 
Unlike the MemNN and DMN architectures, MemN2N can be trained end-to-end and are weakly supervised.
However, the drawback of MemN2N is that it only generates answers 
with one word. 
The proposed LTMN architecture improves over the existing network architectures because  (i) it can be trained end-to-end, (ii) it is weakly supervised, and (iii) can generate answers with multiple words.

%% file: 2-model.tex
\section{Long-Term Memory Networks}
\begin{figure*}[!htb]
 \centering
\includegraphics[width=5.1in]{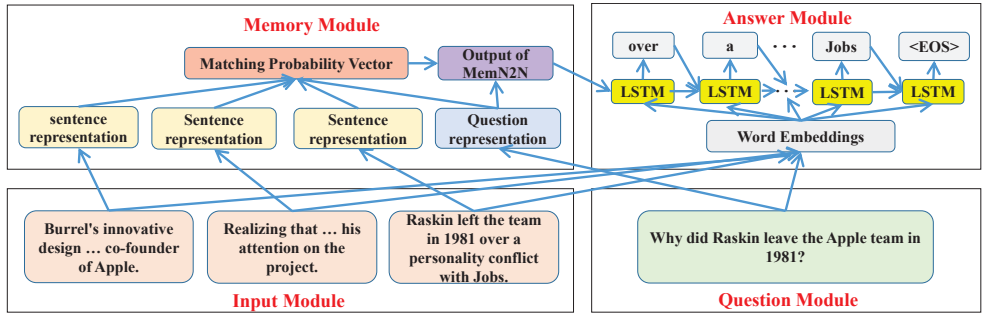}
\caption{The proposed LTMN model.}\label{fig:model}
\end{figure*}


In this section, we describe the proposed Long-Term Memory Network, shown in
Figure \ref{fig:model}.
It includes four modules: input module,
question module, memory module and answer module.
The input module encodes raw text data (i.e., sentences) into
vector representations. Similarly, the question module also encodes
questions into vector representations. The input and question modules can
use the same or different encoding methods. Given the input sentences'
representations, the memory module calculates the matching
probabilities between the question representation and the sentence
representations, and then outputs the weighted sum
of the sentence representations and matching probabilities. Using this weighted sum vector
and the question representation, the answer module
finally generates the answer for the question.

\subsection{Input Module and Question Module}
Let $\{x_i\}_{i=1}^{n}$ represent the set of input sentences.
Each sentence $x_i \in \mathbb{R}^{|V|}$ contains words belonging to a
dictionary $V$, and ends with an end-of-sentence token $<$EOS$>$. 
The goal of the input module is to encode sentences
into vector representations.
The question module, like the input module, aims to encode each question $q \in \mathbb{R}^{|V|}$ into a
vector representation.
Specifically, we use a matrix $A \in \mathbb{R}^{d \times |V|}$ to embed sentences and $B \in \mathbb{R}^{d \times |V|}$ for questions. 

Several methods have been proposed to encode the input sentences or questions. In~\cite{sukhbaatar2015end},
an embedding matrix is employed to embed the sentences in a continuous space
and obtain the vector representations.
\cite{DBLP:conf/icml/KumarIOIBGZPS16,elman1991distributed} use a recurrent neural network
to encode the input
sentences into vector representations.
Our objective is to learn the co-occurrence and sequence relationships between words in the text
in order to generate a coherent sequence of words as answers. 
Thus, we employ a distributed representation learning technique, such as paragraph vectors (paragraph2vec) model~\cite{le2014distributed} to pre-train $A$ and $B$ (with $A=B$) for the real-word SQuAD dataset,
which takes into account the order and semantics among words to encode the input sentences and questions\footnote{We use paragraph2vec in our implementation. Other representation learning mechanisms may be employed in the proposed LTMN model.}.
For synthetic datasets, which are based on a small vocabulary, the embedding matrices $A$ and $B$ are learnt via back-propagation.

\subsection{Memory Module}
The input sentences $\{x_i\}_{i=1}^n$ are embedded using the matrix $A$ as $m_i = A x_i , i=1, 2, \ldots, n; m_i \in \mathbb{R}^d$ and
stored in memory. Note that we use all the
sentences before the question as input, which implies that the proposed model is \textbf{weakly supervised}. 
The question $q$ is also embedded using the matrix $B$ as $u = Bq; u \in \mathbb{R}^d$. 
The memory module then calculates the matching probabilities between the sentences and the question, by
computing the inner product followed by a softmax function as follows:
\begin{equation}
p_i = \mbox{softmax}(u^Tm_i),
\end{equation}
where $\mbox{softmax}(z_i) = e^{z_i} / \sum_j e^{z_j}$.
The probability $p_i$ is expected to be high for all the sentences $x_i$ that are
related to the question $q$.

The output of the memory module is a vector $o \in \mathbb{R}^d$, which can be
represented by the sum over input sentence representations, weighted by the
matching probability vector as follows:
\begin{equation}
o = \sum_i p_i m_i.
\end{equation}

This approach, known as the \emph{soft attention mechanism}, has been used by
\cite{sukhbaatar2015end,bahdanau2015neural}.
The benefit of this approach is that it is easy to compute gradients and
back-propagate through this function.

\subsection{Answer Module}
Based on the output vector $o$ from the memory module and the word
representations from input module, the answer module generates answers for
questions. As our objective is to generate answers with \textbf{multiple words}, we employ
Long Short Term Memory Networks (LSTM) \cite{hochreiter1997long} to generate answers.

The core of the LSTM neural network is a memory unit whose behavior
is controlled by a set of three gates: input, output and forget gates.
The memory unit accumulates the knowledge
from the input data at each time step, based on the values of the gates, and stores this
knowledge in its internal state.
The initial input to the LSTM is the embedding of the begin-of-answer ($<$BOA$>$)
token and its state. We use the output of the memory module $o$, the question
representation $u$, a weight matrix $W^{(o)}$ and bias $b_o$ to generate the
embedding of $<$BOA$>$ $a_0$ as follows:
\begin{equation}
a_0 = \mbox{softmax}(W^{(o)}(o+u) + b_o).
\end{equation}
Using $a_0$ and the initial state $s_0$, LSTM can generate the first word
$w_1$ and its corresponding predicted output $y_1$ and state $s_1$. At each time
step $t$, LSTM takes the embedding of word $w_{t-1}$ and last hidden state
$s_{t-1}$ as input to generate the new word $w_t$.
\begin{equation}
v_t = [w_{t-1}]
\end{equation}
\begin{equation}
i_t = \sigma(W_{iv}v_t + W_{im}y_{t-1} + b_i)
\end{equation}
\begin{equation}
f_t = \sigma(W_{fv}v_t + W_{fm}y_{t-1} + b_f)
\end{equation}
\begin{equation}
o_t = \sigma(W_{ov}v_t + W_{om}y_{t-1} + b_o)
\end{equation}
\begin{equation}
s_t = f_t \odot s_{t-1} + i_t \odot \tanh(W_{sv}v_t + W_{sm}y_{t-1})
\end{equation}
\begin{equation}
y_t = o_t \odot s_t
\end{equation}
\begin{equation}
w_t = \argmax\left[ \mbox{softmax}(W^{(t)}y_t + b_t)\right]
\end{equation}
where $[w_t]$ is the embedding of word $w_t$ learnt from the input module, $\sigma$ and $\odot$ denote the sigmoid function and Hadamard product respectively, and $W^{(t)}$ is a weight matrix and $b_t$ is a bias vector.

The model is trained end-to-end with the loss defined by the cross-entropy between the
true answer and the predicted output
$w_t$, represented using one-hot encoding. The predicted answer is generated by concatenating
all the words generated by the model.

%% file: 4-experiments.tex
\section{Experiments}
In this section, we compare the performance of the proposed LTMN model with the current state-of-the-art 
models for question answering. 
\subsection{Datasets}
We use three datasets: 
the real-world Stanford question answering dataset (SQuAD) \cite{rajpurkar2016squad}, the synthetic single-word answer bAbI 
dataset \cite{weston2016towards}, and the synthetic multi-word answer bAbI dataset, generated by performing vocabulary replacements in the single-word answer bAbI dataset.

\textbf{Stanford Question Answering Dataset (SQuAD)~\cite{rajpurkar2016squad}} contains 100,000+ questions labeled 
by crowd workers on a set of Wikipedia articles. The answer for each question is 
a segment of text from the corresponding paragraph. In order to convert the format of the data to the input format 
of our model (shown in Figure~\ref{fig:example}) , we use NLTK to detect the 
boundary of sentences and assign an index to each sentence and question, in accordance 
with the starting index of the answer provided by the crowd workers. The dataset is thus transformed 
to a question answer dataset containing $18,893$ stories and $69,523$ questions\footnote{The dataset can be downloaded from \url{http://www.acsu.buffalo.edu/~fenglong/}}.
For our experiments, we randomly selected $1,248$ questions for training and 
$1,248$ questions for testing. Each answer contains less than or equal to five words. 

\textbf{The single-word answer bAbI dataset~\cite{weston2016towards}} is a synthetic dataset created to benchmark question answering models. 
It contains $20$ types of question answer tasks, and each task is comprising a set of statements followed by a single-word answer. 
For each question, only some of the statements contain the 
relevant information. The training and test data contains $1,000$ examples for each task.

\textbf{The multi-word answer bAbI dataset.}
As the goal of the proposed model is to generate multi-word answers, we manually generated a new dataset
from the Facebook bAbI dataset, by replacing few words, such as ``bedroom'' and ``bathroom'' 
with ``guest room'', and ``shower room'', respectively. The replacements are listed 
in Table~\ref{tbl:babi-replacements}.

\begin{table}[!htb]
\centering
\caption{Replacements made in the vocabulary of the bAbI dataset to generate the multi-word answer bAbI dataset.}\label{tbl:babi-replacements}
{\small
\begin{tabular}{l|l}
\toprule
Original word & Replacement \\
\midrule
hallway & entrance way\\
bathroom & shower room\\
office & computer science office \\
bedroom & guest room\\
milk&hot water \\
Bill&Bill Gates \\
Fred&Fred Bush \\
Mary&Mary Bush \\
green&bright green \\
yellow&bright yellow \\
hungry&extremely hungry \\
tired&extremely tired \\
\bottomrule
\end{tabular}
}
\end{table}
\subsection{Parameters and Baselines}
We use $10\%$ of the training data for model validation to choose the best parameters.
The best performance was obtained when the learning rate was set to $0.002$, the batch size set to $32$, 
and the weights initialized randomly from 
a Gaussian distribution with zero mean and $0.1$ variance. The model was trained for $200$ epochs.
The paragraph2vec model was set to generate $100$-dimensional 
representations for the input sentences and the questions.

We first compare the performance of the proposed LTMN model with a simple Long Short 
Term Memory network (LSTM) model, as implemented in \cite{sutskever2014sequence} to predict sequences. The LSTM model works by 
reading the story until it comes across a question and outputs an answer, using the information 
obtained from the sentences read so far. Unlike the LTMN model, it does not have an external memory component. We also compare its performance  

On the single-word answer bAbI dataset, we also compare our results with those of the attention based LSTM model (LSTM + Attention)~\cite{hermann2015teaching}, which propagates dependencies between input sentences using an attention mechanism, MemNN~\cite{weston2015memory}, DMN~\cite{DBLP:conf/icml/KumarIOIBGZPS16}, and MemN2N~\cite{sukhbaatar2015end}. These models cannot be applied as-is to the SQuAD and multi-word answer bAbI datasets because they are only capable of generating single-word answers. 

\subsection{Evaluation Measures}
In order to evaluate the performance of all the methods, the following 
measurements are used:
\begin{itemize}
\item Exact Match Accuracy (EMA) represents the ratio of predicted answers which exactly match the true answers.
\item Partial Match Accuracy (PMA) is the ratio of generated answers that 
partially match the correct answers.
\item BLEU score \cite{chen2014systematic}, widely used to evaluate machine translation models, 
measures the quality of the generated answers.
\end{itemize}

\begin{table}[!htb]
\centering
\caption{Test accuracy on the SQuAD dataset.}\label{tab:squad}
{\small
\begin{tabular}{cccccc}
\toprule
Measure & LSTM & LTMN \\
\midrule
EMA & 8.3 & \textbf{10.6} \\
BLEU & 12.4 &  \textbf{17.0}  \\
PMA & 22.8 & \textbf{27.4} \\
\bottomrule
\end{tabular}
}
\end{table}

\subsection{Results}
The performance of the LTMN model is shown in Tables \ref{tab:squad}, \ref{tab:single}, and \ref{tab:multiple} on the 
SQuAD, single-word answer bAbI and multi-word answer bAbI datasets, respectively. 

We observe that LTMN performs better than LSTM in terms of all three evaluation measures, on all the datasets.
On the SQuAD dataset, as the vocabulary is large ($8,969$), the LSTM model cannot learn the embedding matrices 
accurately, leading to its poor performance. However, as the LTMN model employs paragraph2vec, it learns richer  
vector representations of the sentences and questions. In addition, it can memorize and reason over the facts better than 
the simple LSTM model. On the multi-word answer bAbI dataset, the LTMN model is significantly 
better than the LSTM model, especially on tasks 1, 4, 12, 15, 19, and 20. 
The average 
EMA, BLEU, and PMA scores of LTMN are about $30\%$ higher than those of the LSTM model. 
The single-word answer bAbI dataset's vocabulary is small (about $20$), so 
we learn the embedding matrices $A$ and $B$ using back-propagation, instead of using paragraph2vec to obtain the vector 
representations. In Table~\ref{tab:single}, we observe that the LTMN model achieves accuracy close to the strongly 
supervised MemNN and DMN models on $4$ out of the $20$ bAbI tasks, despite being weakly supervised, and achieves better accuracy than the weakly-supervised LSTM+Attention and MemN2N on $7$ tasks. 
The proposed LTMN model also offers the additional capability of generating multi-word answers, unlike these baseline models. 
%
%

\begin{table*}[!htb]
\centering
\caption{Test accuracy (EMA) on the single-word answer bAbI dataset}\label{tab:single}
{\small
\begin{tabular}{lcccccc}
\toprule
\multirow{2}{*}{Task} & \multicolumn{4}{c}{Weakly Supervised}   & 
\multicolumn{2}{c}{Strongly Supervised} \\
\cmidrule(r){2-5} \cmidrule(r){6-7} 
 & LSTM & LSTM + Attention & MemN2N &  LTMN & MemNN & DMN \\
\midrule
1: Single Supporting Fact &  50 & 98.1& 96 & \textbf{98.2} & 100 & 100 \\
2: Two Supporting Facts & 20 &33.6&  61 & 41.6 & 100 & 98.2 \\
3: Three Supporting Facts & 20 &25.5& 30 & 23.8  &  100 & 95.2 \\
4: Two Argument Relations & 61&98.5 & 93 & 98.1 & 100 & 100  \\
5: Three Argument Relations & 70&97.8 & 81 &79.5  & 98 & 99.3 \\
6: Yes/No Questions & 48 &55.6& 72 & \textbf{81.8} & 100 & 100 \\
7: Counting & 49 &80.0& 80 & \textbf{80.2} & 85 & 96.9 \\
8: Lists/Sets & 45 &92.1& 77 & 72.6 & 91 & 96.5 \\
9: Simple Negation & 64 &64.3&72  & 65.4  & 100 & 100 \\
10: Indefinite Knowledge & 46 &57.2&63  &\textbf{87.0}  & 98 & 97.5 \\
11: Basic Coreference & 62 &94.4& 89 &84.7  & 100 & 99.9 \\
12: Conjunction & 74 &93.6& 92 & \textbf{97.9} & 100 & 100 \\
13: Compound Coreference & 94& 94.4 &  93& 90.3 & 100 & 99.8 \\
14: Time Reasoning & 27 & 75.3 & 76  & 74.3 & 99 & 100 \\
15: Basic Deduction & 21 &57.6& 100 &\textbf{100}  & 100 & 100 \\
16: Basic Induction & 23 &50.4& 46 & 43.5 &  100&  99.4\\
17: Positional Reasoning &51& 63.1&57 & 57.0 & 65 & 59.6 \\
18: Size Reasoning &  52& 92.7& 90& 90.7 & 95 & 95.3 \\
19: Path Finding &8  &11.5 & 9  & 11.4 &36  & 34.5 \\
20: Agent's Motivations & 91 &98.0& 100 &  \textbf{100}& 100 & 100 \\ \midrule
Mean (\%) & 48.8 & 71.7& 73.9 & 73.9 & 93.4 & 93.6 \\
\bottomrule
\end{tabular}
}
\end{table*}

\begin{table*}[!htb]
\centering
\caption{Test accuracy on the multi-word answer bAbI dataset.}\label{tab:multiple}
{\small
\begin{tabular}{lcccccccc}
\toprule
\multirow{2}{*}{Task} & \multicolumn{3}{c}{LSTM}   & \multicolumn{3}{c}{LTMN} \\
\cmidrule(r){2-4} \cmidrule(r){5-7} 
 & EMA & BLEU &  PMA & EMA & BLEU &  PMA\\
\midrule
1: Single Supporting Fact & 36.5 & 38.8 &41.1  & \textbf{97.0} & \textbf{97.2} & \textbf{97.3} \\
2: Two Supporting Facts & 26.6 & 29.7 & 32.7  & 31.3 & 34.5 &37.6 \\
3: Three Supporting Facts & 17.1 & 20.3 & 23.6 & 24.5 &27.2  &29.8\\
4: Two Argument Relations & 48.2 & 50.1 &  51.9 & \textbf{97.9} & \textbf{98.0} & \textbf{98.0} \\
5: Three Argument Relations & 45.3 & 49.3 &53.2 & 77.9 & 80.1 &82.2 \\
6: Yes/No Questions & 53.8 & 53.8 &53.8 & 66.1 & 66.1 & 66.1\\
7: Counting & 69.5 & 69.5 & 69.5  & 78.4 & 78.4 & 78.4\\
8: Lists/Sets & 62.1 & 66.7 & 71.8 &  82.1 & 85.6 & 89.3\\
9: Simple Negation & 57.4 & 57.4 &57.4 & 69.2 & 69.2 & 69.2 \\
10: Indefinite Knowledge & 44.4 & 44.4 &44.4 & 84.7 &84.7  & 84.7 \\
11: Basic Coreference & 33.1 & 35.1 &37.0 & 83.3 & 83.7 & 84.0\\
12: Conjunction & 33.1&35.7 &38.2 & \textbf{99.3} & \textbf{99.3} & \textbf{99.4} \\
13: Compound Coreference &33.6  &35.8 & 37.9 & 87.7 & 88.5 & 89.2 \\
14: Time Reasoning & 24.6 &24.6  &24.6  & 74.4 & 74.4 & 74.4 \\
15: Basic Deduction & 46.4 & 46.4 & 46.4 & \textbf{100} & \textbf{100}& \textbf{100} \\
16: Basic Induction & 46.8 & 51.6 &  56.3& 42.4 & 47.0 &51.6 \\
17: Positional Reasoning & 55.1 & 55.1 & 55.1 & 55.5 & 55.5 &55.5 \\
18: Size Reasoning & 51.9 & 51.9 & 51.9 & 89.6 & 89.6 & 89.6 \\
19: Path Finding & 8.1 & 35.1 & 56.4 & 11.3 & 59.1 & \textbf{100} \\
20: Agent's Motivations & 83.3 & 84.6 & 85.3 & \textbf{100} & \textbf{100} & \textbf{100} \\ \midrule
Mean (\%) & 42.2 & 46.8 & 49.4 & 72.6 & 75.9 & 78.8 \\
\bottomrule
\end{tabular}
}
\end{table*}

%% file: 5-conclusion.tex
\section{Conclusions}
Question answering is an important and challenging task in 
natural language processing. Traditional 
question answering approaches are simple query-based approaches, which cannot memorize and reason over the input text. Deep neural networks with memory have been employed to alleviate this challenge in the literature. 

In this paper, we proposed the Long-Term Memory Network, a novel 
recurrent neural network, which can encode raw 
text information (the input sentences and questions) into vector 
representations, form memories, find relevant information in the input sentences to answer the questions, and 
finally generate multi-word answers using a long short term memory network. The 
proposed architecture is a weakly supervised model and can be trained end-to-end. 
Experiments on both synthetic and real-world datasets demonstrate the remarkable 
performance of the proposed architecture.

In our experiments on the bAbI question \& answering tasks, 
we found that the proposed model fails to perform as well as the completely supervised memory networks 
on certain tasks. In addition, the model performs poorly when the input sentences are very long and the vocabulary is large, as it cannot calculate 
the supporting facts efficiently. In the future, we plan to expand the model to handle long 
input sentences, and improve the performance of the proposed network.
